# Application of Confidence Intervals to the Autonomous Acquisition of High-level Spatial Knowledge

Lambert E. Wixson

Computer Science Department
University of Rochester
Rochester, NY 14627
wixson@cs.rochester.edu

## Abstract

Objects in the world usually appear in context, participating in spatial relationships and interactions that are predictable and expected. Knowledge of these contexts can be used in the task of using a mobile camera to search for a specified object in a room. We call this the object search task. This paper is concerned with representing this knowledge in a manner facilitating its application to object search while at the same time lending itself to autonomous learning by a robot. The ability for the robot to learn such knowledge without supervision is crucial due to the vast number of possible relationships that can exist for any given set of objects. Moreover, since a robot will not have an infinite amount of time to learn, it must be able to determine an order in which to look for possible relationships so as to maximize the rate at which new knowledge is gained. In effect, there must be a "focus of interest" operator that allows the robot to choose which examples are likely to convey the most new information and should be examined first.

This paper demonstrates how a representation based on statistical confidence intervals allows the construction of a system that achieves the above goals. An algorithm, based on the Highest Impact First heuristic, is presented as a means for providing a "focus of interest" with which to control the learning process, and examples are given.

## 1 Introduction

Objects in the world usually appear in context, participating in spatial relationships and interactions that are predictable and expected. Recently, we have begun to investigate the task of *object search* [Wixson and Ballard, 1989; Wixson, 1990b]. The task is simply stated: Using a mobile camera, find a specified object that is somewhere in the room. The search should reliably find the desired object while minimizing the time spent looking. One obvious step towards achieving this goal is to use high-level knowledge to predict likely locations of objects and to order the camera gazes[1] considered so as to minimize the time to find the object. For example, when looking for a light switch humans immediately look near doorways. When looking for a pen, we first look on desks. Introspection reveals that much of daily life can be characterized as object search, and that much of object search is driven by knowledge about the relationships that commonly hold between objects. [Garvey, 1976] referred to this use of high-level knowledge as *indirect search*.

Given that knowledge about spatial contexts can be very useful to an object search system, how should the system acquire this knowledge? It is unreasonable to expect human programmers to encode it all for several reasons. First, there are so many common contexts that we take for granted in everyday life that it is unlikely that they all could be listed by humans, let alone entered into an object search system. Second, situations are bound to arise that result in changes to the stored knowledge. For example, contextual knowledge about the locations of workstations might change from one academic department to another: in Computer Science departments the likelihood of there being a workstation being on top of a randomly-chosen desk might be substantially larger than in Biology departments. This leads to the conclusion that a robot must be able to learn this high-level knowledge on its own.

---

[1] A camera gaze is a 3-D position and orientation.

Several considerations arise in building a mechanism by which a robot can autonomously wander around its environment and acquire this knowledge. First, the learning is bandwidth-limited. Since relationships must be detected visually and since a robot is likely to have only one camera platform, only one (or at most, a very small number) of relationships can be examined at any one time. For example, if the robot has somehow decided to examine the relationships in which a certain chair is participating, it probably cannot at the same time be examining the relationships in which a certain coffee cup is participating. The robot will not have an infinite amount of time to acquire context knowledge. It is therefore important to have a mechanism that can be used to decide the order in which relationships should be examined. For example, given that the robot knows the locations of a coffee cup and a chair, how should it decide which object's relations to examine first? One fundamental factor in this decision must be the degree of knowledge about the relationships in which these objects commonly appear. If the system has seen many examples of chairs, but few examples of coffee cups, then it seems intuitive that the robot should look at the coffee cup's relations first. To make such a decision, a representation of context is needed that allows the characterization and formalization of knowledge and/or uncertainty.

This paper demonstrates how a representation based on statistical confidence intervals allows the construction of a system that supports such reasoning. It starts with a discussion of the nature of the relationships to be characterized. Following this, Section 3 discusses the use of these relationships in object search and motivates the probability measurements that an object search system must acquire. Unfortunately, these measurements are numerous and a robot with limited time must determine which are the most important and should be acquired first. Sections 4 and 5 present our answer to this problem. Section 4 shows how instances in which these relationships hold or do not hold can be viewed as Bernoulli trials; this insight allows us to characterize the probability that the relationship holds as an interval whose width is is easily computable. Section 5 describes how a type of Highest Impact First heuristic [Feldman and Yakimovsky, 1974; Weber, 1989; Chou, 1988; Swain et al., 1989] that measures the size of the intervals can be used by the system to acquire knowledge in order of importance.

Before starting, a brief word about terminology and notation is necessary. For our purposes, any object in the world is an instance of exactly one object type.[2] To denote a specific object, we will use a lower-case identifier. To denote an object type, we will use an identifier whose first letter is capitalized. In addition, the type of an object $obj$ can be denoted as $type(obj)$. Thus, for example, an object denoted $desk5$ might be an instance of the $Desk$ object type, and if so, then $type(desk5) = Desk$.

## 2 Relationships

What are the properties that a representation of spatial relationships should have for our purposes? As stated previously, the system must be able to learn about the relationships between objects from scratch. This means that the spatial relationships must be computable from visual data. In addition, it seems doubtful that detailed metric information about relationships is necessary; information that, for example, objects of type $X$ are always .7 meters to the right of objects of type $Y$, will probably not be helpful for two reasons. First, objects in the real world are unlikely to always be in metrically uniform relationships. Additionally, a camera can easily view a relatively large volume in a single image and hence such exact information about where to point the camera is unnecessary.

This section begins with a discussion of related work on characterizing relations among objects. Following this, a general representation for arbitrary relations, based on volumes emanating from each object, is presented.

### 2.1 Related Work

Rumelhart et al.[1986] studied the related problem of how schemata could be formed for rooms. Specifically, the goal was to show how knowledge about the presence or absence of objects could excite or inhibit other hypotheses about the presence of other objects in the room. Thus, they were studying only one relation, which might be called IN-THE-SAME-ROOM. Unfortunately, however, this relation is not sufficient for the purposes of the object search task, since our goal is to constrain the locations of objects.

---

[2] For this paper we assume that objects have only one type, although there do not appear to be large obstacles to using a type hierarchy in the future.

Computer vision researchers and linguists have studied the spatial prepositions that are used to describe spatial layouts [Talmy, 1983; Takahashi et al., 1989]. The major observations from these studies have been that the primitives we use to characterize relationships are almost all pairwise, characterizing one object's spatial disposition in terms of another's. In addition, many relationships are described using an object-centered partial coordinate system. For example, one might say "The bike is in front of the house.", thereby using the house to project a volume in which one can say the bike is contained. This coordinate system may be partial in the sense that only certain relationships may be useful with a specified object. For instance, soda cans have a well-defined top and bottom, but not a front, back, left side, or right side.

## 2.2 A general representation for object-centered spatial relations

The observations that many useful relations are pairwise, object-specific, and take advantage of an object-centered coordinate system, combined with the ability of mobile cameras to inspect a volume of space, suggests a representation in which relations correspond to object-specific object-centered volumes. More specifically, a set of relation names (such as TOP-OF-SODA-CAN or NEAR-SODA-CAN) is associated with each object model. (We adopt the convention that the name of the model must appear as the suffix of each of the names associated with the model, resulting in the convenience that each relation name is unique.) For a given object type $B$, this set of relation names will be denoted $RELATIONS(B)$. Each relation in $RELATIONS(B)$ is associated with a volume, defined with respect to the model's coordinate system. For a given object type $B$, the set of these volumes will be denoted $VOLS(B)$. Given an object $b$ of type $B$, the volume (in world coordinates) associated with relation $rel \in RELATIONS(B)$ will be denoted $v_{rel}(b)$. Given an object $obj$, an object $b$ of type $B$, and a relation $rel \in RELATIONS(B)$, we say that $rel(obj, b)$ holds iff $obj$ lies at least partially within $v_{rel}(b)$.

This object-specific representation of volumes provides great flexibility. Many objects, such as soda cans, might only have a few relations defined for them (for example, RELATIONS(SODA-CAN) might contain only ABOVE-SODA-CAN, BELOW-SODA-CAN, and NEAR-SODA-CAN) while others, such as desks, might have many relations (such as LEFT-OF-DESK, RIGHT-OF-DESK, IN-FRONT-OF-DESK, BEHIND-DESK, ABOVE-DESK, UNDER-DESK, etc.). Contact relations (such as TOUCHING-BOTTOM-OF-SODA-CAN, for instance) can be approximated using very thin volumes. It is the task of the programmer to choose the relations and define the corresponding volumes for each object. It should be noted also that given the location and orientation of two objects A and B, it is relatively straightforward to compute the truth of each of these relations or to determine any new camera positions necessary for this computation.

## 3 Application to Indirect Search

There are several ways in which the probabilistic representation of the high-level knowledge described above can guide a system performing object search. One method, if the poses[3] of certain objects are known, is to use these locations to predict likely locations for the desired object. For example, suppose the location of a desk is known in advance. If the robot is then told to search for a chair, it could start by searching near the desk. Moreover, if the robot knows there is also a typewriter located next to the desk, it should start its search by examining the space which is near both the typewriter and the desk. We call this the *location constraint* method.

A second method, called *detectability-driven search*, which requires no initial knowledge of poses, determines whether there are objects that are more easily detectable than the desired object (perhaps due to color, size, or distinguishing geometric features) and if they can be found, are likely to provide information about the relative location of the target object. If so, then these objects are searched for, and once these are found then location constraint may be applied. For example, when searching for a small object such as a pencil, it may be advantageous to first search for a larger object which is likely to constrain the location of the pencil, such as a desk. The system constructed by Garvey [1976], did this to a limited extent - it could compute a score for a plan that tried to find the object directly and it could score plans for finding the object indirectly. It should be noted, however, that plans for indirect object detection were hardwired, preprogrammed strategies.

---

[3]The *pose* of an object is its location and orientation.

In addition, Garvey's system constructed no indirect search plans based purely on its probabilistic data about spatial relationships, nor did it attempt to learn these probabilities autonomously.

These search methods were considered [Wixson, 1990a] in order to determine the form of the context knowledge necessary for their execution. Consideration of possible implementations of these methods led to the hypothesis that context knowledge can be usefully represented by simple estimates of the probability that, given an object $a$ of type $A$, there exists at least one object $b$ of type $B$ such that $rel(b,a)$ holds, for $rel \in RELATIONS(A)$. We denote this probability as $P(rel(B))$. (Note that the notation $P(rel(B))$ implicitly specifies the type $A$, since the type is the suffix of the relation name $rel$.)

Thus, for example, if the system can identify cups and tables, and can identify the following relations for each,

- RELATIONS(Cup) = {NEAR-CUP, TOUCHING-BOTTOM-OF-CUP, INSIDE-CUP}

- RELATIONS(Table) = {NEAR-TABLE, TOUCHING-TOP-OF-TABLE, UNDER-TABLE, ABOVE-TABLE}

then it would store information such as $P(\text{NEAR}-\text{CUP}(\text{Table}))$ and $P(\text{UNDER}-\text{TABLE}(\text{Cup}))$. $P(\text{NEAR}-\text{CUP}(\text{Table}))$, for example, might be defined in probabilistic first-order logic as

$$P(\ \exists xy.\text{Cup}(x) \wedge \text{Table}(y) \wedge \text{NEAR}-\text{CUP}(y,x)\ |$$
$$\exists x.\text{Cup}(x)).$$

In general, given a set of object types $TYPES$, one must keep track of $\sum_{t \in TYPES}(\|TYPES\| \times \|RELATIONS(t)\|)$ probabilities. Since the size of $TYPES$ is the dominating term, this sum is $O(T^2)$, where $T = \|TYPES\|$.

[Wixson, 1990a] also concluded that either point-valued or interval-valued measures of $P(rel(B))$ could be used to drive the location-constraint and detectability-driven search strategies. The subsequent sections of this paper show how statistical confidence intervals may be used to represent this knowledge and that these intervals facilitate the autonomous learning of this knowledge by a robot.

## 4 Representation

This section describes the use of statistical confidence intervals to represent $P(rel(B))$. As will be shown in Section 5, besides being easily represented and computed, the changes in the size of the confidence intervals as more examples are seen can be used to direct the knowledge acquisition process.

The relations developed in Section 2.2 are binary-valued. Given the location and orientation of a single object $a$, then for each $rel \in RELATIONS(type(a))$, for every physical object $obj$ in the environment, $rel(obj, a)$ either holds or does not hold.

Statisticians refer to experiments which produce binary-valued results (either "success" or "failure") as Bernoulli trials. Given $n$ independent Bernoulli trials, resulting in a total of $y$ successes, and a parameter $0 < \alpha < 1$, one can construct an interval which, with probability $1 - \alpha$, contains the unknown chance of success $p$. This interval can be closely approximated [Larsen and Marx, 1981, pp.228-232] by $(p_0(y,n) - d(y,n), p_0(y,n) + d(y,n))$, where

$$p_0(y,n) = \frac{\frac{y}{n} + \frac{z_{\alpha/2}^2}{2n}}{1 + \frac{z_{\alpha/2}^2}{n}} \qquad (1)$$

$$d(y,n) = \frac{\frac{z_{\alpha/2}}{\sqrt{n}}\sqrt{(\frac{y}{n})(1 - \frac{y}{n}) + \frac{z_{\alpha/2}^2}{4n}}}{1 + \frac{z_{\alpha/2}^2}{n}} \qquad (2)$$

and $z_{\alpha/2}$ is chosen such that if $Z$ has a normal distribution, $P(Z \geq z_{\alpha/2}) = \alpha/2$.

The width of this interval is determined by the function $d$. The size of the interval can be thought of as representing the precision of one's knowledge about the probability of success on any given trial.

We can treat the perception of $rel(obj, a)$ for particular objects $obj$ and $a$ (of types $Obj$ and $A$, respectively) and relation $rel \in RELATIONS(A)$ as a Bernoulli trial, and hence can construct a confidence interval estimate of $P(rel(Obj))$. This interval can be computed given that we have observed relation $rel$ to hold in $y_{rel,Obj}$ cases out of the $n_A$ observations of an object of type $A$. The confidence interval containing $P(rel(Obj))$ with probability $1 - \alpha$, for a fixed $\alpha$ (typically .10), will be denoted as $INT(rel, Obj)$.

# 5 Autonomous Acquisition of Spatial Contexts

In this section we describe how the interval representation described above can be used to provide a "focus of interest" for acquiring new information. Assume that we are given a vision system that can:

- Recognize any object from a set of object types, *TYPES*, where each object type is associated with the volumetric information needed to determine the truth of the relations in which the object can participate.

- Given the location and orientation (in world coordinates) of an object *obj* of type $t$, $t \in TYPES$, can determine which areas of the scene (in world coordinates) correspond to the volumes associated with *obj*.

Though the first is still a research problem, these assumptions are reasonable. The first simply requires that objects can be recognized and that their poses can be estimated in order to compute the truth of spatial relations, and the second relies only on the ability to compute some geometric transforms. The reliance on a world coordinate frame may be undesirable, but it is certainly possible to impose some sort of coordinate system on the local environment.

We now want to "turn the system loose", to allow it to autonomously examine its environment, acquiring knowledge about relationships between objects. Such knowledge acquisition might occur in a training period where the robot's task is simply to acquire data, or perhaps during times where the robot has no other tasks to handle. There remains the question, however, of what objects and relationships the robot should concentrate on during the acquisition process. At any given time, there will be a set of objects (called **UNEXAMINED**) that the system has recognized in the scene but whose relations have not been analyzed. A mechanism is required that allows the system to choose the order in which these objects and the relations in which they participate should be examined. This is an issue because the robot does not have infinite time to explore its environment. Time is limited and the robot must try to maximize its information gain at all times. Hence, the unexamined objects should be examined in order of increasing potential information. That is, the object about whose type the least is known should be examined first, and the object about whose type the most is known should be examined last.

```
procedure initialize-context-knowledge:
(1)      foreach t1 ∈ TYPES
(2)          n_t1 := 0
(3)          foreach rel ∈ RELATIONS(t1)
(4)              foreach t2 ∈ TYPES
(5)                  y_{rel,t2} := 0
(6)                  INT(rel,t2) := [0, 1]
```

Figure 1: Procedure to initialize spatial knowledge

In order to construct such an order in which objects should be investigated, it is necessary to have a method for measuring our knowledge about the spatial contexts in which each object type appears. Such a measure can be constructed by considering the role of the function $d$ that determines the width of the confidence intervals that represent our knowledge about spatial contexts. Given a set of $n$ trials with $y$ successes, one can compute the maximum possible change in the width, $impact(y, n)$, if one more experiment is conducted:

$$impact(y, n) = 2 * \max(\ |d(y,n) - d(y, n+1)|, \\ |d(y,n) - d(y+1, n+1)|\ ) \quad (3)$$

The algorithm presented in Figures 1 and 2 uses the *impact* function to provide a measure of knowledge. It is based on the idea that whenever there is a choice of objects to investigate, the choice should be made so as to cause the biggest possible changes in the intervals stored in the database. In other words, the object with the largest impact should be investigated first.

Figure 1 contains the procedure for initializing the system before any learning has taken place. As described in Section 4,

- $n_{T1}$ is the number of objects of type $T1$ which have been observed.

- $y_{rel,T2}$, where $rel \in RELATIONS(T1)$ for exactly one object type $T1$, is the number of times that, given an object $a$ of type $T1$, there has existed at least one object $b$ of type $T2$ such that $rel(b, a)$ holds.

- $INT(rel, T2)$ is the confidence interval estimate of $P(rel(T2))$.

At initialization time, no objects have been recognized yet, so there is no data about the spatial con-

```
procedure learn-spatial-contexts:
(1)     UNEXAMINED := ∅
(2)     EXAMINED := ∅

(3)     while mode = ACQUIRE-SPATIAL-KNOWLEDGE do:
(4)         if UNEXAMINED = ∅ then:
(5)             Using the mobile camera, find some nearby object obj which
                    is recognizable and is not in EXAMINED
(6)             If no such object can be found, then halt.
(7)             Add obj to UNEXAMINED

(8)         UNEXAMINED-TYPES := {type(u) | u ∈ UNEXAMINED}

(9)         foreach ut ∈ UNEXAMINED-TYPES
(10)            foreach rel ∈ RELATIONS(ut)
```
(11) $\quad d_{ut} := average\ impact(y_{rel,t}, n_{ut})\ over\ all\ t \in TYPES$

(12) *Choose a type* S ∈ UNEXAMINED-TYPES *such that* $d_S = \max_{ut} d_{ut}$
(13) *Choose an object* obj ∈ UNEXAMINED *such that* $type(\text{obj}) = S$
(14) Add obj *to* EXAMINED

(15) RELATED-OBJECTS := {i | i *is an object in the scene and*
     $\exists r \in RELATIONS(S)\ such\ that\ r(i, \text{obj})\ holds$ }
(16) RELATED-TYPES := {< r, type(i) > | i *is an object in the scene and*
     $r \in RELATIONS(S)\ and\ r(i, \text{obj})\ holds$ }

(17) *increment* $n_S$
(18) foreach <r, t> ∈ RELATED-TYPES
(19) $\quad$ *increment* $y_{r,t}$
(20) $\quad INT(r,t) := (p_0(y_{r,t}, n_S) - d(y_{r,t}, n_S),$
     $\qquad\qquad\qquad p_0(y_{r,t}, n_S) + d(y_{r,t}, n_S))$

(21) UNEXAMINED := (UNEXAMINED ⋃ RELATED-OBJECTS) - EXAMINED

Figure 2: Algorithm for autonomous knowledge acquisition

texts of objects. The no-data state is represented by the [0, 1] confidence interval.

Figure 2 contains the acquisition procedure, which would be invoked whenever the robot commences knowledge acquisition. The UNEXAMINED variable will hold the set of objects that the system has recognized in the scene but whose relations have not yet been examined. EXAMINED will be the set of objects in the scene whose relations have already been examined by the system. Lines 1 and 2, since no objects have been examined yet, initialize UNEXAMINED and EXAMINED to the null set.

Line 8 stores the set of the types of objects in UNEXAMINED in UNEXAMINED-TYPES.

Lines 9 - 13 are the crucial portion of this algorithm. They choose which object's context is to be examined next. For each type ut, $d_{ut}$ is an estimate of the average impact that examination of a new instance of type ut will have. The impact is measured by the simple *impact* function described by Equation 4. The impacts are averaged to produce $d_{ut}$. An object of the type with maximum $d_{ut}$ is chosen to be examined. [4]

Lines 15 and 16 reflect the result of deploying the sensors so as to examine all the relationships in which obj participates. The spatial context of obj is examined by attempting to identify all objects that can be related to obj when obj is considered to be the reference object. More formally, our goal is to update $INT(r,t)$, for every $r \in RELATIONS(S)$ and every $t \in TYPES$. To do this, the robot will inspect the spatial volumes which correspond to all the possible relations that an object could have to the reference object obj. It attempts to recognize every object i that intersects one of these volumes, and if it is recognizable, adds it to the RELATED-OBJECTS set and adds information about the relation and object i's type to RELATED-TYPES.

Lines 17 - 20 perform the database updating given the set of related objects. First, the count of the number of objects of type S is incremented. Next, the y-value associated with the pair composed of each relation and $type(i)$ is incremented if necessary, and finally the intervals are updated.

Finally, in line 21, the related objects, since they have now been identified, are added to UNEXAMINED. Any objects that have already been examined are then removed from UNEXAMINED.[5]

The algorithm described above provides us with an unsupervised "interest" measure that a robot can use to direct its learning of spatial relationships. This is valuable since humans obviously cannot decide in advance on a fixed set of attention-focusing procedures. An extended example of this algorithm can be found in [Wixson, 1990a]. Confidence intervals have also been used recently by Kaebling [1990] to guide experimentation for agents learning action strategies.

## 6 Conclusion

This paper has described a technique by which a robot can represent knowledge about spatial contexts by simply counting the number of observed cases where two objects were in a certain primitive relation. This information can be used to compute confidence intervals for the actual probability that a certain relationship holds. An algorithm was presented that used the sizes of these confidence intervals to judge what objects in the environment are most interesting (from the standpoint of learning spatial contexts) at any given time. In addition, an application of the confidence intervals to the location constraint method of object search was presented.

Let us now consider the computational feasibility of the suggested representation. A system with human-level recognition capabilities must be able to recognize 30,000 different types of objects[Biederman, 1985]. Since our method requires $O(T^2)$ storage for $T$ objects, it would require drastic modification to handle an object database of this size. However, for robot systems with under 1000 object models, an implementation of our method using sparse matrices to store the y-values certainly seems feasible.

There is one key aspect of the acquisition algorithm in need of further work. This has to do with the temporal aspects of selecting objects from which to learn. More specifically, if a certain object is examined at time $t$, how much time must elapse before the system can learn from the object again? For example, suppose that yesterday the object search system examined the relationships in which desk5 participated in order to learn more about which objects are commonly found on desks. Is it acceptable for desk5 to be examined again today, since it is likely that the objects on top of it may have changed?

---

[4] Note that embellishments could be made to this framework in order to take into account varying costs of examining each object.

[5] Note that this last step involves some kind of world representation that allows us to store which objects have been examined and which have not.

Besides this problem, there are several obvious areas in which this work can be extended. First, objects should be allowed to be members of a type hierarchy rather than of a single type. Thus, if the system has learned that books can often be found on desks and shelves, and desks and shelves all are derived from the horizontal-object class, then it could infer that books are often found on horizontal objects. Other, more theoretical, extensions might be to analyze this approach in terms of information theory, and to investigate the use of fuzzy sets in order to use more than just {true, false} values for the relationships.

## Acknowledgments

Henry Kyburg provided the idea of using intervals to represent the probabilities of relations. Dana Ballard, Chris Brown, Randal Nelson, Michael Swain, Cesar Quiroz, and Nat Martin provided valuable advice and interesting discussions.

# Session 5:

## Belief Network Decomposition